\documentclass{article}

\usepackage[accepted]{_ICML/icml2022}

\usepackage[utf8]{inputenc} %
\usepackage[T1]{fontenc}    %
\usepackage[hidelinks]{hyperref}       %
\usepackage{url}            %
\usepackage{booktabs}       %
\usepackage{amsfonts}       %
\usepackage{nicefrac}       %
\usepackage{microtype}      %
\usepackage{xcolor}         %

\usepackage{graphicx}

\usepackage{amssymb, mathtools, amsmath}

\usepackage{amssymb, mathtools, amsmath}
\usepackage{tcolorbox}
\usepackage{xcolor}

\usepackage{microtype}
\usepackage{graphicx}
\usepackage{booktabs} %
\usepackage[font=small]{caption}
\usepackage[inline]{enumitem}

\usepackage[colorinlistoftodos]{todonotes}

\usepackage{wrapfig}
\usepackage{enumitem}
\usepackage{multicol}

\usepackage{subfig}
\usepackage[font=small,labelfont=bf]{caption}

\usepackage{pifont}

\usepackage{algorithm}
\usepackage{algorithmic}

\hypersetup{
    colorlinks = true,
    citecolor=blue,
    linkcolor=blue,
    urlcolor=blue
}

\usepackage{enumitem}
\setlist{nolistsep,leftmargin=*}
\definecolor{darkgreen}{rgb}{0.0, 0.5, 0.0}
\definecolor{blue}{rgb}{0.0, 0.47, 0.75}
\definecolor{dartmouthgreen}{rgb}{0.05, 0.5, 0.06}
\definecolor{drab}{rgb}{0.59, 0.44, 0.09}
\definecolor{navyblue}{rgb}{0.0, 0.0, 0.5}

\newcommand{\mb}[1]{\mathbf{#1}}

\newcommand{\E}{\mathbb{E}}
\newcommand{\KL}{\mathbb{KL}}
\newcommand{\prodl}{\prod^{L-1}_{l=1}}

\newcommand{\vb}{\,\vert\,}

\newlist{todolist}{itemize}{2}
\setlist[todolist]{label=$\square$}
\newcommand{\cmark}{\ding{51}}%
\newcommand{\xmark}{\ding{55}}%

\icmltitlerunning{}

\begin{document}

\twocolumn[

\icmltitle{SCHA-VAE: Hierarchical Context Aggregation for Few-Shot Generation}

\icmlsetsymbol{equal}{*}

\begin{icmlauthorlist}
\icmlauthor{Giorgio Giannone}{to}
\icmlauthor{Ole Winther}{to,goo}
\end{icmlauthorlist}

\icmlaffiliation{to}{Technical University of Denmark}
\icmlaffiliation{goo}{University of Copenhagen}

\icmlcorrespondingauthor{Giorgio Giannone}{gigi@dtu.dk}

\icmlkeywords{Few-Shot Generation, Hierarchical Inference, Aggregation}

\vskip 0.3in
]

\printAffiliationsAndNotice{} %

\begin{abstract}
A few-shot generative model should be able to generate data from a novel distribution by only observing a limited set of examples. 
In few-shot learning the model is trained on data from many sets from distributions sharing some underlying properties such as sets of characters from different alphabets or objects from different categories.
We extend current latent variable models for sets to a fully hierarchical approach with an attention-based point to set-level aggregation and call our method SCHA-VAE for Set-Context-Hierarchical-Aggregation Variational Autoencoder. 
We explore likelihood-based model comparison, iterative data sampling, and adaptation-free out-of-distribution generalization. 
Our results show that the hierarchical formulation better captures the intrinsic variability within the sets in the small data regime.
This work generalizes deep latent variable approaches to few-shot learning, taking a step toward large-scale few-shot generation with a formulation that readily works with current state-of-the-art deep generative models.
\end{abstract}

\section{Introduction}

Humans are exceptional few-shot learners able to grasp concepts and function of objects never encountered before~\cite{lake2011one}. 
This is because we build internal models of the world so we can combine our prior knowledge about object appearance and function to make well educated inferences from very little data~\cite{tenenbaum1999bayesian,lake2017building, ullman2020bayesian}. 
In contrast, traditional machine learning systems have to be trained tabula rasa and therefore need orders of magnitude more data.
In the landmark paper on modern few-shot learning~\citet{lake2015human} demonstrated that with a strong prior hand-written symbols from different alphabets can be generated few-shot and distinguished one-shot, i.e.~when a letter is shown for the first time.

Few-shot learning~\cite{vinyals2016matching, snell2017prototypical, finn2017model} and related approaches aiming at learning from little labelled data at test time~\cite{schaul2010metalearning, hospedales2020meta} have recently gained new interest thanks to modeling advances, availability of large diverse datasets and computational resources. 
Building efficient learning systems that can adapt at inference time is a prerequisite to deploy such systems in realistic settings. 
Much attention has been devoted to \emph{supervised few-shot learning}. 
The problem is typically cast in terms of an adaptive conditional task, where a small support set is used to condition explicitly~\cite{garnelo2018conditional} or implicitly~\cite{finn2017model} a learner, with the goal to perform well on a query set. 
The high-level idea is to train the model with a large number of small sets, and inject in the model the capacity to adapt to new concepts from few-samples at test time.

\begin{figure}[ht]
    \centering
    \includegraphics[width=.9\linewidth]{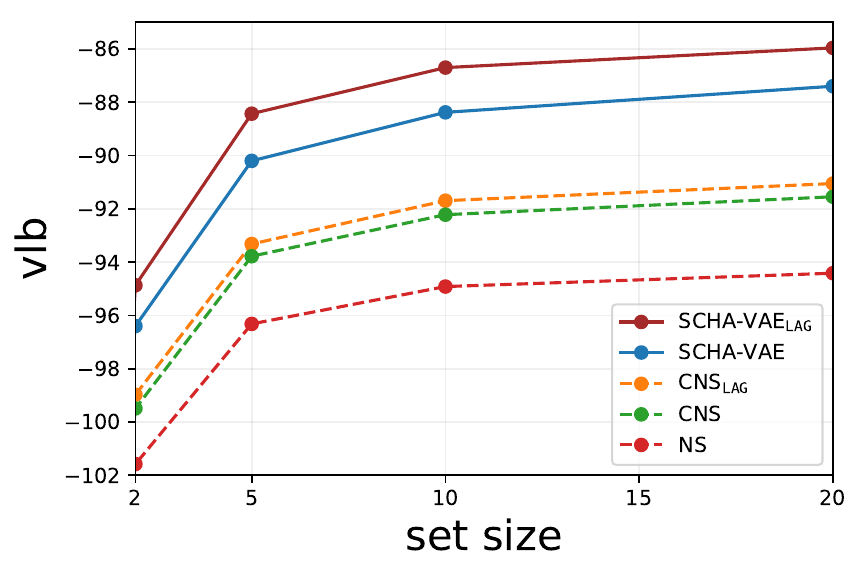}
    \caption{Lower bounds for NS-style models trained with input set size 5 varying the set cardinality from 2 to 20 on Omniglot test set.
    The convolutional latent space (CNS) is fundamental for performance improvement, increasing the expressivity of the context representation.
    Hierarchical inference over $\mathbf{c}$ (SCHA-VAE) and learnable aggregation (LAG) improve the generative performance in a monotonic way, providing evidence that SCHA-VAE can effectively adapt to different input set size outperforming the NS baselines. 
    }
    \label{fig:main_figure}
\end{figure}
Comparatively less work has been developed on \emph{few-shot adaptation in generative models}~\cite{edwards2016towards, reed2017few, bartunov2018few}.
This is partially because of the challenging nature of learning joint distributions in an unsupervised way from few-samples and difficulties in evaluating such models.
Few-shot generation has been limited to simple tasks, shallow and handcrafted conditioning mechanisms and as pretraining for supervised few-shot learning. 
Consequently there is a lack of quantitative evaluation and progress in the field.
\begin{figure}
    \centering
    \includegraphics[width=.9\linewidth]{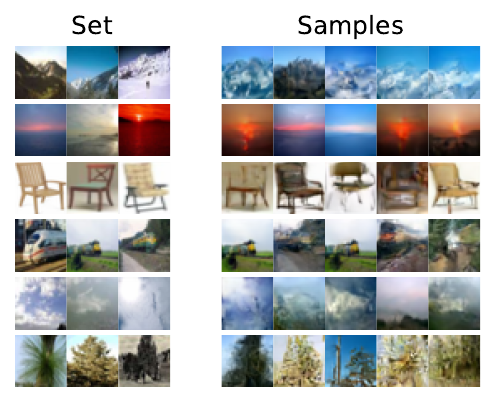}
    \caption{Visualization for a few-shot generative model trained on FS-CIFAR100. 
    Given a small set (between 2 and 20) of samples from a class, our model can extract information from complex realistic images, aggregate the content/class information, and generate samples that are consistent with the conditioning set.}
    \label{fig:main_set}
\end{figure}
In this work we aim to solve these issues for few-shot generation in latent variable models. 
This class of models is promising because provides a principle way to include adaptive conditioning using latent variables.
The setting we consider is that of learning from a large quantity of homogeneous sets, where each set is an un-ordered collections of samples of one concept or class. 
At test time, the model will be provided with sets of concepts never encountered during training and sets of different cardinality.
We consider explicit conditioning in a hierarchical formulation, where \emph{global latent variables carry information about the set at hand}.
A pooling mechanism aggregates information from the set using a hard-coded (mean, max) or learned (attention) operator.
The conditioning mechanism is implemented in a shallow or hierarchical way: the hierarchical approach helps to learn better representations for the input set and gradually merges global and local information between the aggregated set and samples in the set.
To handle input sets of any size the mechanism has to be permutation invariant. 
The conditional hierarchical model can naturally represent a family of generative models, each specified by a different conditioning set-level latent variable. 
Learning a full distribution over the input set increases the flexibility of the model.

\paragraph{Contribution.} %
\begin{itemize}
\item %
We increase the input set representation expressivity of latent variable models for sets through \textbf{hierarchical inference} and a \textbf{learnable aggregation} mechanism. 
\item We explore forward and iterative \textbf{sampling strategies} for the marginal and the predictive distributions implicitly defined by few-shot generative models.  
\item We study the few-shot \textbf{transfer} for this new class of models, exploring generalization to set cardinality, new classes and new datasets, providing evidence that a hierarchical set representation increases the expressivity of few-shot generative models.
\end{itemize}

\section{Generative Models over Sets}

In this section we present the modeling background for the proposed few-shot generative models.
The Neural Statistician 
(NS,~\cite{edwards2016towards}) is a latent variable model for few-shot learning.
Based on this model, many other approaches have been developed~\cite{garnelo2018neural, bartunov2018few, hewitt2018variational}.
The NS is a hierarchical model where two collections of latent variable are learned: a \emph{task-specific} summary statistic $c$ with prior $p(c)$ and a \emph{per-sample} latent variable $\mb{z}$ with prior $p(\mb{z}| c)$:
\begin{equation}
p(X) = 
\int p({c}) \prod_{s=1}^S 
\left[\int p(x_s | \mb{z}_s, {c}) p(\mb{z}_s | {c}) d\mb{z}_s \right] d {c} \ ,
\end{equation}
where $X=\{x_1,\ldots,x_S\}$ assuming the data set size is $S$ and $p(\mb{z} | {c})$ in general is a hierarchy of latent variables~\cite{sonderby2016ladder, maaloe2019biva}: 
$p(\mb{z}|{c}) = p(z_L | {c})\prod^{L-1}_{l=1} p(z_l | z_{l+1}, {c})$ with $\mb{z}=\{{z}_1,\ldots,{z}_L\}$.

In the original NS model, the authors factorize the lower-bound wrt $c$ and $\mb{z}$ as: 
\begin{equation}
q({c}, \mb{Z}  | X) = q({c} | X) \prod_{s=1}^S q(\mb{z}_s | {c}, x_s),
\end{equation}
where $\mb{Z}=\{\mb{z}_1,\ldots,\mb{z}_S\}$ and $q(\mb{z}_s | x_s, \mb{c})$ is in general a hierarchy of latent variables.
In the NS, the moments of the conditioning distribution over $c$ are computed using a simple sum/average based formulation 
$
r = \sum^S_{s=1} h(x_s),
\label{eq:h_simple_aggr}
$;
and then $r$ is used to condition $q(c | r)$.
This idea is simple and straightforward, is permutation invariant, enabling efficient learning and a clean lower-bound factorization. 
But it has a limitation: 
with such formulation the model expressivity and capacity to extract information from the set are limited.
The pooling mechanism assumes strong homogeneity in the context set and the generative process.
However, the model formulation opens the door for more advanced invariant aggregations based on attention~\cite{vaswani2017attention} and graph approaches~\cite{velickovic2017graph}).

In the next section, we move beyond simple aggregation to learn more expressive few-shot generative models able to deal with variety and complexity in the conditioning set. We propose a hierarchical merging of information between conditioning $\mb{c}$ and sample $\mb{z}$; and a learnable agggregation mechanism for the context set.

\section{Hierarchical Few-Shot Generative Models} 
\label{sec:SCHA-VAE}
\begin{figure*}
    \centering
    \includegraphics[width=0.4\textwidth]{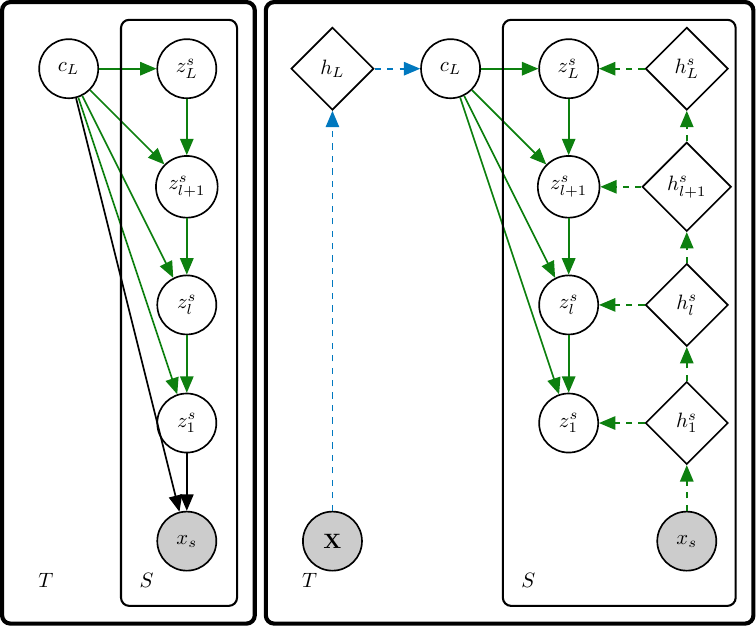}
    \quad\quad
    \includegraphics[width=0.4\textwidth]{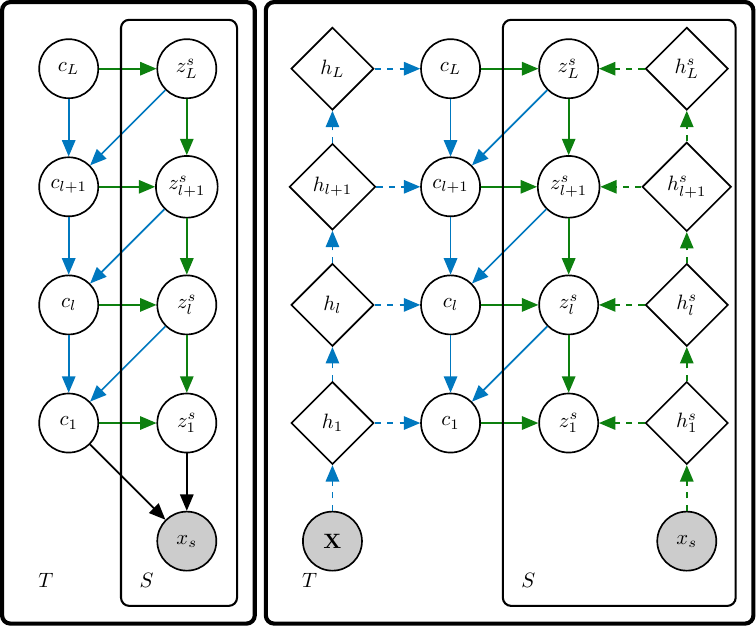}
    \caption{Generation and inference for a Neural Statistician (left) and a Hierarchical Few-Shot Generative Model (right). 
    The generative model is composed by two collections of hierarchical latent variables, $\mb{c}$ for the sets $X=\{x_s\}^{S}_{s=1}$ and $\mb{z}_s$ for the samples $x_s$. 
    The generative process is repeated $S$ times and the full model is run on $T$ different sets or tasks.
    The two variables are learned jointly, increasing expressivity and improving sampling.
    The generative and inference models can share parameters.}
    \label{fig:gen}
\end{figure*}
Our goal is to learn a generative model over sets, i.e. unordered collections of observations, able to generalize to new datasets given few samples.
In this model, $\mb{c}$ is a collection of latent variables that represent a set $X$. 
We learn a posterior over $\mb{z}$ conditioned on $\mb{c}$, able to generate samples accordingly to elements $x$ in the input set.
The model has to be expressive: 
I) \emph{hierarchical} - to increase the functions that the model can represent and improve the joint merging of set-level information $\mb{c}$ and sample information $\mb{z}$, and 
II) \emph{flexible} - to handle input sets of any size and complexity, improving the way the model extracts and organizes information in the conditioning set.
A fundamental difference between our proposal and previous models is the intrinsically hierarchical inference procedure over $\mb{c}$ and $\mb{z}$. 

\paragraph{Generative Model.}
The generative model factorizes the joint distribution $p(X, \mb{Z}, \mb{c})$. In particular can be written as:
\begin{equation}
p(X | \mb{Z}, \mb{c})
p(c_L)\left[
p(Z_L | c_L)
\prodl
p(Z_l, c_l | Z_{l+1}, c_{l+1})
\right] 
\end{equation}
where $Z_l=\{z^1_{l},\ldots,z^S_{l}\}$ are latent for layer $l$ in the model and sample $s$ in the input set $X=\{x_s\}^{S}_{s=1}$; $c_l$ is a latent for layer $l$ in the model and the input set $X$. 
We factorize the likelihood and prior terms over the set as: 
\begin{align}
p(X | \mb{Z}, \mb{c}) &= \prod^S_{s=1} p(x_s | \mb{z}_s, \mb{c}) \\ 
p(Z_l, c_l | Z_{l+1}, c_{l+1}) &= 
\prod^S_{s=1} p(z^{s}_{l} | z^{s}_{l+1}, c_l)
~p(c_l | c_{l+1}, Z_{l+1}).
\end{align}
In this formulation the context $p(c_l | c_{l+1}, Z_{l+1})$ is a distribution of the previous context representation $c_{l+1}$ and the previous latent representation for the samples $Z_{l+1}$ as illustrated in Figure~\ref{fig:gen}.

\paragraph{Approximate Inference.}
Learning in the model is achieved by amortized variational inference~\cite{jordan1999introduction, hoffman2013stochastic}.
The hierarchical formulation leverages structured mean-field approximation to increase inference flexibility.
The approximate posterior parameterizes a joint distribution between latent representation for context and samples.
We factorize $q(\mb{c}, \mb{Z} | X)$ following the generative model:
\begin{equation}
q(c_L | X) 
\left[ q(Z_L | c_L, X)
\prod^{L-1}_{l=1} q(Z_l, c_l | Z_{l+1}, c_{l+1}, X) 
\right] 
\end{equation}
where we factorize the posterior using a top-down inference formulation~\cite{sonderby2016ladder}, 
merging top-down stochastic inference with bottom-up deterministic inference from the data.
We factorize the posterior terms over the set as:
\begin{align}
\begin{split}
&q(Z_l, c_l | Z_{l+1}, c_{l+1}, X) = \\
&\prod^S_{s=1}
q(z^{s}_{l} | z^{s}_{l+1}, c_l, x_s)
~q(c_l | c_{l+1}, Z_{l+1}, X).
\end{split}
\end{align}
\paragraph{Lower bound.}
In latent variable models~\cite{rezende2014stochastic, kingma2013auto} we maximize a \emph{per-sample} lowerbound over a large dataset.
In few-shot generative models we maximize a lower-bound over a large collection of \emph{small} sets. 
This detail is important because, even if the dataset of sets is iid by construction, learning in the few-shot scenario relies explicitly on common structure within a small set.
For example, each small set can be an unordered collection of observations for a specific class or concept, like a character or face; and we rely on common structure between these observations using aggregation and conditioning on $\mb{c}$.
The variational lower bound for $\log p(X)$ is obtained using the variational distribution $q=q(\mb{c}, \mb{Z} | X)$:
\begin{align}
\label{eq:ELBO}
\begin{split}
~ &\E_{q}
\Bigg[ \sum^S_{s=1} \log p(x_s | \mb{z}_s, \mb{c}) \Bigg] + \\
&\E_{q} \Bigg[ \sum^{L-1}_{l=1} \sum_{s=1}^S \log \frac{p(z^s_l | z^s_{l+1}, c_l)}{q(z^s_l | z^s_{l+1}, c_l, x_s)} + \sum_{s=1}^S \log \frac{p(z^s_L | c_L)}{q(z^s_L | c_L, x_s)} \Bigg] + \\
~& 
\E_{q} \Bigg[ \sum^{L-1}_{l=1} \log \frac{p(c_l | c_{l+1}, Z_{l+1})}{q(c_l | c_{l+1}, Z_{l+1}, X)} \Bigg] -\KL(q(c_L | X), p(c_L))
\ .
\end{split}
\end{align}

The lower-bound can be split in three main components: an expected log likelihood term, divergences over $\mathbf{Z}$ and over $\mathbf{c}$.
The final per-sample loss for $T$ sets of size $S$ then is $\mathcal{L} = \mathbb{E}_{p(\mathcal{X}_S)} l(X) = 1/N \sum^T_{t=1} l(X_t)$, where $l(X_t)$ is the negated lower-bound for set $X_t$ and $N=T\,S$.

\paragraph{Hierarchical Set Representation.}
The core idea in Eq.~\ref{eq:ELBO} is that a hierarchical representation for the input set improves few-shot generation. Indeed adding a hierarchy over $c$ we largely increase the model capacities. 
I) The hierarchical formulation for $p(c_l | c_{l+1}, Z_{l+1})$ and $q(c_l | c_{l+1}, Z_{l+1}, X)$ increases the variational posterior flexibility, and hierarchical inference has been effective to scale per-sample latent variable models to large domains~\cite{maaloe2019biva, vahdat2020NVAE, child2020very}. Following this reasoning, a hierarchical representation for $c$ should increase the model expressivity when dealing with sets.
II) We can aggregate context set information at multiple scales and resolution by design, merging a generic $c_l$ with information extracted from previous resolutions. 
Doing so the model can iteratively refine the class information contained in the set, an essential capacity when dealing with few-samples from a new class. 
III) The hierarchical formulation gives us also a principle way to share per sample information $z$ and per-set information $c$ at different level of abstraction in the model, again increasing the model expressivity. 
IV) On a practical level, the hierarchical formulation can easily be coupled with modern large deep latent variable models~\cite{child2020very}, without the risk of forgetting context information increasing depth or coarse granularity in the context, providing a promising direction for large scale few-shot generative architectures.

\subsection{Sampling}
\begin{algorithm}
\small
   \caption{Iterative Sampling}
   \label{alg:sampling}
\begin{algorithmic}
   \STATE {\bfseries Input:} set $X$
   \STATE $c_L \sim q(c_L|X)$
   \STATE $\mb{z},\mb{c}_{<L}\sim p(\mb{z},\mb{c}_{<L}|c_L)$
   \STATE $x\sim p( x|\mb{z},\mb{c})$
   \REPEAT
   \STATE $\tilde{X}=[X,x]$
   \STATE $\tilde{Z},\mb{c}\sim q(\tilde{Z},\mb{c}|\tilde{X})$
   \STATE $\tilde{X}' \sim p(\tilde{X}'|\tilde{Z},\mb{c})$
   \STATE $x = x'$
   \UNTIL{\texttt{Converged}}
   \RETURN $x$
\end{algorithmic}
\end{algorithm}
We may be interested in either sampling unconditionally $x\sim p(x)$ or from the predictive distribution $x\sim p(x|X)$. Unconditional sampling may be performed exactly using the generative model as illustrated in Figure~\ref{fig:gen}: first sample the hierarchical prior $\mb{z},\mb{c} \sim p(\mb{z},\mb{c})$ and then sample the likelihood $x\sim p(x|\mb{z},\mb{c})$. 
Conditional sampling $x\sim p(x|X)$ can be done approximately using the variational posterior $q(\mb{Z},\mb{c}|X)$ as a replacement for the exact posterior $p(\mb{Z},\mb{c}|X)$ as outlined in Algorithm \ref{alg:sampling}. In the single pass approach (adapted from the NS) a sample from: 
\begin{equation}\label{eq:pred_dist_q_C_L_|_X}
    \int p( x|\mb{z},\mb{c}) p(\mb{z},\mb{c}_{<L}|c_L)q(c_L|X) d\mb{z}d\mb{c} \approx p(x|X)  
\end{equation}
is generated.
This approach is not ideal because $\mb{c}_{<L}$ is only modeled jointly with the latent $\mb{z}$ for the new sample while omitting the latent $\mb{Z}$ for $X$. We can introduce this dependence in a Markov chain approach adapted from the missing data imputation framework proposed in Appendix F of \cite{rezende2014stochastic}.
We augment $X$ with the sample $x$ generated in the single pass approach: $\tilde{X}=[X,x]$. From this we can construct a distribution $\hat{p}(\tilde{Z},\mb{c}|\tilde{X})$, where $\tilde{Z}=[Z,z]$ is the corresponding augmented latent. From this distribution and the likelihood we can construct a transition kernel:
\begin{equation*}
     \hat{p}(x'|x,X) =  \int p(\tilde{X}'|\tilde{Z},\mb{c}) \hat{p}(\tilde{Z},\mb{c}|\tilde{X}) dX' d\tilde{Z} d\mb{c} \ . 
\end{equation*}
If $\hat{p}(\tilde{Z},\mb{c}|\tilde{X})=p(\tilde{Z},\mb{c}|\tilde{X})$ then we can show that $p(x|X)$ is an eigen-distribution for the transition kernel and thus sampling the transition kernel will under mild conditions converge to a sample from $p(x|X)$.
There are several ways to construct $\hat{p}(\tilde{Z},\mb{c}|\tilde{X})$ from the hierarchical variational and prior distributions. One possibility is shown in Algorithm \ref{alg:sampling}. Alternatively, one may 
sample $c_L,\tilde{Z}_L \sim q(c_L,\tilde{Z}_L|\tilde{X})$, generate the remainder of the latent from the prior hierarchy and new samples $X',x'$ from the likelihood. %
If the variational is exact, these approaches are equivalent and exact.
In the experimental section we report results for the different approaches.

\subsection{Learnable aggregation (LAG)}
\begin{algorithm}
\small
   \caption{Learnable Aggregation (LAG)}
   \label{alg:lag}
\begin{algorithmic}
   \STATE {\bfseries Input:} set $H=\{h_s = h(x_s)\}^{S}_{s=1}$
   \STATE base aggregation $r=1/S \sum^{S}_{s=1} h_s$
   \STATE compute query, key, value: $q = q(r)$, $k_s = k(h_s)$, $v_s = v(h_s)$
   \STATE weights $\alpha(q, k_s) = \sigma(\texttt{dot}(q, k_s))$ 
   \STATE {\bfseries aggregate}
   \STATE $r_{\texttt{LAG}} = \sum^{S}_{s=1} \alpha(q, k_s)~v_s$
   \STATE {\bfseries sample}
   \STATE $c \sim q(c ~|~ r_{\texttt{LAG}})$
\end{algorithmic}
\end{algorithm}

A central idea in few-shot generative models is to condition the generative model with a permutation invariant representation of the input set.
For a NS such operator is a hard-coded per-set statistic.
This approach~\cite{edwards2016towards, garnelo2018neural} maps each sample in $X$ independently using $\{h(x_s)\}^S_{s=1}$ and then aggregating $r_L = \sum^S_{s=1} h(x_s)$ to generate the moments for $q(c_L | r_L)$.
This idea is simple and effective when using homogeneous and small sets for conditioning. 
Another choice is a relation network~\cite{sung2018learning, rusu2018meta} followed by aggregation.
\begin{wrapfigure}{r}{0.4\columnwidth}
\centering
    \includegraphics[width=\linewidth]{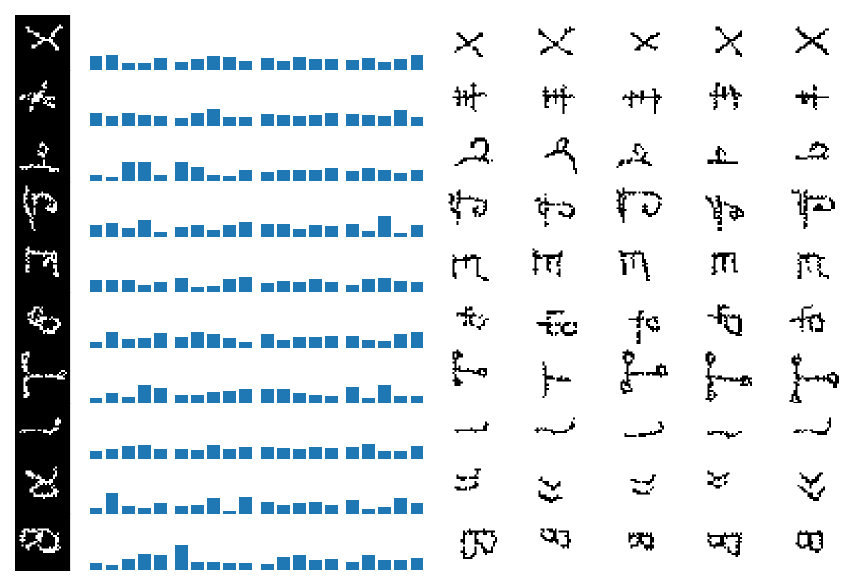}
\caption{Attention. Sample formation in a Convolutional Neural Statistician with learnable aggregation. For each sample (black background) we plot the attention bars over the input sets (in white background) for four different heads.}
\label{fig:attention_omniglot}
\end{wrapfigure}
However, the adaptation capacities of the model are a function of how we represent the input set and a more expressive learnable aggregation mechanism can be useful.
In the general scenario, a few-shot generative model should be able to extract information from any conditioning set $X$ in terms of variety and size.
In this paper we consider a multi-head attention-based learnable aggregation (LAG, Algorithm~\ref{alg:lag}) inspired by~\cite{lee2019set} that can be used in each block of the hierarchy over $c$ (Figure~\ref{fig:attention_omniglot}).
Using LAG we can account for statistical dependencies in the set, handle variety between samples in the set and generalize better to input set dimensions. 
In the experimental section we provide extensive empirical analysis of how using LAG improves the generative and transfer capacity of the model. More details in Appendix~\ref{appendixA}.

\section{Experiments}
In this section we discuss experimental setup and results for our model.
In particular for all the models our interests are: 
I) Quantitative evaluation of few-shot generation;
II) Few-shot conditional and unconditional sampling from the model.
III) Transfer of generative capacities to new datasets and input set size.
IV) Evidence that the hierarchical context representation is useful for the model.

\paragraph{Datasets.} 
We train models on binarized Omniglot~\cite{lake2015human}, CelebA~\cite{liu2015faceattributes} and FS-CIFAR100~\cite{oreshkin2018tadam}.
For Omniglot we consider two scenarios: the train-test split (also known as background-evaluation split) proposed in~\cite{lake2015human}: in this scenario all the characters at test time are from new alphabets. 
And the variant proposed in~\cite{edwards2016towards}, where all the characters are mixed together and randomly assigned to train and test splits.
We perform quantitative evaluation on Omniglot, MNIST~\cite{lecun1998mnist}, DOUBLE-MNIST~\cite{mulitdigitmnist}, TRIPLE-MNIST~\cite{mulitdigitmnist}, CelebA and FS-CIFAR100.
We resize all the binary datasets to 28x28, CIFAR100 to 32x32, and CelebA to 64x64.

\paragraph{Training Details.}
We follow the approach proposed in~\cite{edwards2016towards} for training with sets.
We create a large collection of small sets, where each set contains all or some of the occurrences for a specific class or concept in a dataset: a character for Omniglot; the face of an identity for CelebA.
Both datasets contain thousands of characters/identities with an average number of 20 occurrences per class. 
For this reason they are a natural choice for few-shot generation.
Then we split the sets in train/val/test sets.
Using these splits we can dynamically create sets of different dimensions, generating a new collection of training sets at each epoch during training.
For training we use the episodic approach proposed in~\cite{vinyals2016matching}.
Each input set is a homogeneous (one concept or class) collection of samples. 
During training the input set size is always between 2 and 20.
The NS architecture is a close approximation of~\cite{edwards2016towards}.
We describe additional details in Appendix~\ref{appendixC}.

\paragraph{Baselines.}
We use a VAE - which does not explore set information - and the Neural Statistician (NS) as baselines. 
We consider two main model variants.
Such variants are characterized by different design choices for the conditioning mechanism:
standard aggregation mechanism (MEAN pooling) and learnable ones (LAG).
We also consider a Convolutional Neural Statistician (CNS) where the latent space is shaped with convolutions at a given resolution.
\emph{We call our approach Set-Context-Hierarchical-Aggregation Variational Autoencoder} (SCHA-VAE, pronounced shave) where an additional hierarchy over $\mathbf{c}$ is employed.

\paragraph{Model Design.}
NS-based models are difficult to scale and performance tends to plateau increasing depth. 
This can be explained considering that $c\sim q(c | X)$ has to be shared through all the network and information can easily be lost in the stochastic sampling process.
We overcome these challenges and stabilize SCHA-VAE when increasing depth, we use a shared routing path between $c$ and $z$: we learn bias term $b^c$ and $b^z$ for the top layer, and merge all sampling information in these paths: $b^{c}_l = b^{c}_{l+1} + f(X, b^{z}_{l+1}) + c_{l}$ and $b^{z}_l = b^z_{l+1} + f(x, b^{c}_l) + z_{l}$, where $z_l$ and $c_l$ are samples from layer $l$. 
We then share information through these path, helping to stabilize and preserve information through the hierarchy.
We additionally remove any form of normalization (batch/layer norm) in the hierarchical formulation. This is not directly related to performance, but to challenge the model to perform pure few-shot generation at train and test time, without relying on batch statistics.
Our implementation is inspired by~\cite{child2020very}.

\subsection{Test log-likelihood}
With these experiments we test the generalization properties of the model on few-shot generation.
We explore the behavior of the model increasing the input size.
Evaluation of generative properties is performed using the log likelihood lower bound (ELBO) and approximating the log marginal likelihood with 1000 importance samples.
In Table~\ref{tab:generative_metrics_omniglot} and~\ref{tab:generative_metrics_in_out_distro} we consider few-shot generalization on Omniglot and MNIST with disjoint classes and set size 5.
In Figure~\ref{fig:main_figure} and~\ref{fig:generative_metrics_omniglot_cardinality} we show the models behavior increasing the input set cardinality at test time. All the models are trained with input sets of size 5 on Omniglot. Then at test time we vary the size of the context set between 2 and 20.
\begin{table*}
\small
\begin{center}
\caption{Generalization on disjoint Omniglot classes trained on set size 5 for
    a VAE, NSs with mean/learnable aggregation (MEAN/LAG),
    convolutional variants (C) and for a SCHA-VAE with a hierarchy over c.
    We report minus the 1-sample lower bound (NELBO), and minus the 1k importance sample lower bound (NLL).
    $\textbf{z}_{1:L}$: hierarchical representation for sample.
    $\textbf{c}_{1:L}$: hierarchical representation for set.
    NS based models process the sets at one resolution without hierarchical inference over c.
    SCHA-VAE processes the sets at multiple resolutions with hierarchical inference over c.
    We see that the hierarchical set representation with agggregation at multiple resolutions is essential to improve generative performance.}
    \setlength\tabcolsep{4.0pt}
    \begin{tabular}{lcccccccccc}
    \toprule
    & & & &\multicolumn{2}{c}{Omniglot-ns} & \multicolumn{2}{c}{Omniglot-back-eval} & \multicolumn{1}{c}{MNIST}
    & \\
    & Agg & $\textbf{z}_{1:L}$ & $\textbf{c}_{1:L}$ & $\downarrow$ NELBO & $\downarrow$ NLL (1k) & $\downarrow$ NELBO & $\downarrow$ NLL (1k) & $\downarrow$ NELBO & Params (M) \\
    \midrule
    VAE & $-$ & \cmark & $-$ & 101.5 $\pm$ .1 & 95.9& 101.4 $\pm$ .1 & 88.4 & 124.7 $\pm$ .1 & $13.3$ \\
    NS  & MEAN & \cmark & \xmark & 96.6 $\pm$ .1 & 92.4 & 91.9 $\pm$ .1  & 85.6 & 120.3 $\pm$ .1 & $14.7$ \\
    CNS & MEAN & \cmark & \xmark & 92.9 $\pm$ .1  & 89.7 & 83.4 $\pm$ .1 & 77.5 & 118.2 $\pm$ .2 & $16.9$ \\
    CNS & LAG  & \cmark & \xmark & 92.8 $\pm$ .1  & 89.7 & 83.8 $\pm$ .1  & 77.6 & 117.6 $\pm$ .1 & $17.2$ \\
    SCHA-VAE (ours) & MEAN & \cmark & \cmark & 89.4 $\pm$ .1 & \textbf{85.8}& 79.4 $\pm$ .1 & 72.7 & \textbf{115.3} $\pm$ .2 & $11.9$ \\
    \textbf{SCHA-VAE (ours)} & LAG & \cmark & \cmark & \textbf{88.4} $\pm$ .3  & \textbf{85.4}& \textbf{77.9} $\pm$ .3 & \textbf{71.5} & \textbf{114.9} $\pm$ .1 & $12.8$ \\
    \bottomrule
    \end{tabular}
\label{tab:generative_metrics_omniglot}
\end{center}
\end{table*}

\begin{table*}
\small
\begin{center}
\caption{Generalization on Omniglot for different set size and augmentation scheme. We train one model on sets of size 5 and test on standard and augmented (translation, rotation, flipping) Omniglot data. 
For the in in-distro we test on \emph{known-classes}; out-distro on \emph{unknown-classes}. We use minus the 1-sample lower bound (NELBO) to evaluate the models. Lower is better. }
\setlength\tabcolsep{4.0pt}
    \begin{tabular}{lcccc|cccc|cccc|cccc}
    \toprule
    & \multicolumn{8}{c}{Standard} 
    & \multicolumn{8}{c}{Augmented} \\
    & \multicolumn{4}{c}{in-distro}
    & \multicolumn{4}{c}{out-distro}
    & \multicolumn{4}{c}{in-distro}
    & \multicolumn{4}{c}{out-distro} \\
    & 2& 5& 10& 20 
    & 2& 5& 10& 20 
    & 2& 5& 10& 20
    & 2& 5& 10& 20 \\
    \midrule
    NS      & 96.8& 91.2& 89.6& 88.8& 101.5 & 96.6 & 95.1 & 94.7 & 107.8& 102.6& 101.1& 100.5& 111.1& 106.1 &104.6 & 103.9\\
    CNS     & 97.8 & 91.7 & 90.0 & 89.2 & 98.8& 92.8& 91.2& 90.6 & 107.7& 101.7& 100.0& 99.4& 108.6& 102.7& 100.9 &100.1\\
    \textbf{SCHA-VAE}  & \textbf{91.5} & \textbf{85.2} & \textbf{83.5} & \textbf{82.7} & \textbf{96.2}& \textbf{89.7}& \textbf{87.9} & \textbf{87.0} & \textbf{103.0}& \textbf{96.6}& \textbf{94.8}& \textbf{94.1}& \textbf{105.9}& \textbf{99.2}& \textbf{97.2} &\textbf{96.2}\\
    \bottomrule
    \end{tabular}
\label{tab:generative_metrics_in_out_distro}
\end{center}
\end{table*}
The NS is the original Neural Statistician with mean aggregation.
CNS employs a convolutional set representation.
The models with LAG use an attention-based learnable aggregation, that builds an adaptive aggregation mechanism for each set.
For SCHA-VAE we use the same two aggregation mechanisms.
SCHA-VAE improves performance in terms of ELBO and likelihood on different Omniglot test splits and transfer to MNIST without adaptation better than the baselines.
The hierarchical formulation for $\mathbf{c}$ provides most of the improvement. Adding the learnable aggregation mechanism gives an additional boost.
Increasing the set (2-20), SCHA-VAE can aggregate more and more information, learning a better model for the data under different regimes (in-distro, out-distro) and augmentation schemes.
See Appendix, Table~\ref{tab:autoregressive} for additional quantitative evaluation with few-shot generative models proposed in the literature, where we compare SCHA-VAE with autoregressive-based models for few-shot generation~\cite{bartunov2018few, hewitt2018variational}.

\subsection{Sampling}
In latent variable models like VAEs we can only perform unconditional sampling.
In a NS we have different ways of sampling as explained in Section 3.1.
In particular two main sampling approaches can be used:
I) conditional sampling, where we sample the predictive distribution $p(x | X)$ relying on the inference model.
II) unconditional sampling, sampling $p(X)$ through $c \sim p(c)$ and then $p(z | c)$.
In Figure~\ref{fig:sampling_omniglot} we show
(from left to right) stochastic reconstructions, input sets, conditional samples, and unconditional samples.
\begin{figure}[ht]
    \centering
    \includegraphics[width=\linewidth]{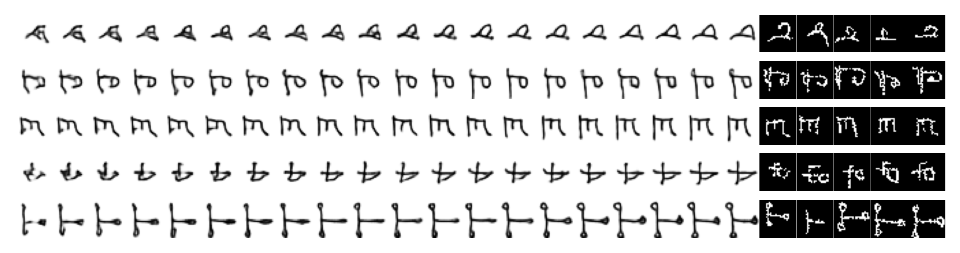}
    
    \includegraphics[width=\linewidth]{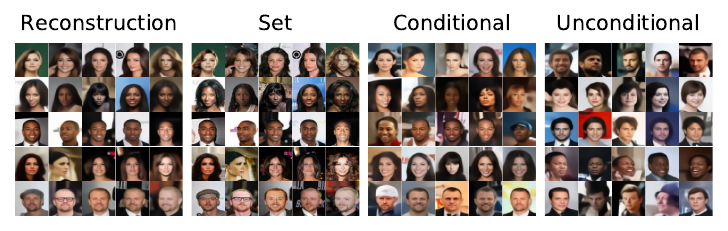}
    \caption{Sampling. Different ways to sample the model.
    (Top): Refined samples obtained using Algorithm~\ref{alg:sampling}. 
    Given a small set from an unknown character (right on black background), we sample the model and then refine iteratively using the inference model. We show 20 iterations from left to right. 
    We can see how the generative process refines its guess at each iteration improving $\mathbf{c}$ and $z$ in a joint manner. \\
    (Bottom):
    Stochastic reconstruction, input sets, conditional sampling, and unconditional sampling (sometimes referred to as imagination).
    The models are trained on subsets of CelebA and tested on disjoint identities.
    Appendix~\ref{appendixB} for larger version.
    }
    \label{fig:sampling_omniglot}
\end{figure}
For simple characters there is almost no difference between conditional and refinement sampling. However, when the model is challenged with a new complex character (Figure~\ref{fig:sampling_omniglot}, top), the refinement procedure greatly improves the adaptation capacities and visual quality of the generated samples.
In the rightmost column in the bottom part of Figure~\ref{fig:sampling_omniglot}  we see fully unconditional samples. The model, given a set representation $\mathbf{c}$, generates consistent symbols, corroborating the assumption that the model learns a different distribution for each $\mathbf{c}$, greatly increasing the model flexibility and representation capacities. 

\subsection{Transfer}
With these experiments we explore few-shot generation in the context of transfer learning.
We use the same models we trained on Omniglot and we test on unseen classes in a different dataset. We use MNIST test set (10 classes), DOUBLE-MNIST test set (20 classes) and TRIPLE-MNIST test set (200 classes). 
The datasets increase in complexity and in size.
We expect relatively good performance on the simple one and worse performance on the more complex one. We resize all the datasets to 28x28 pixels using BOX resizing.
In Figure~\ref{fig:generative_metrics_transfer_cardinality} we report likelihoods transfer increasing the input set size on the three datasets.
Our models perform better on all three datasets. 
Attention-based aggregation is essential for good performance on few-shot transfer.
Again our models perform better than the baselines and attention-based aggregation is important for good performance on concepts from different datasets.
\begin{figure*}[ht]
  \centering
  \subfloat[MNIST]{\includegraphics[width=.3\linewidth]{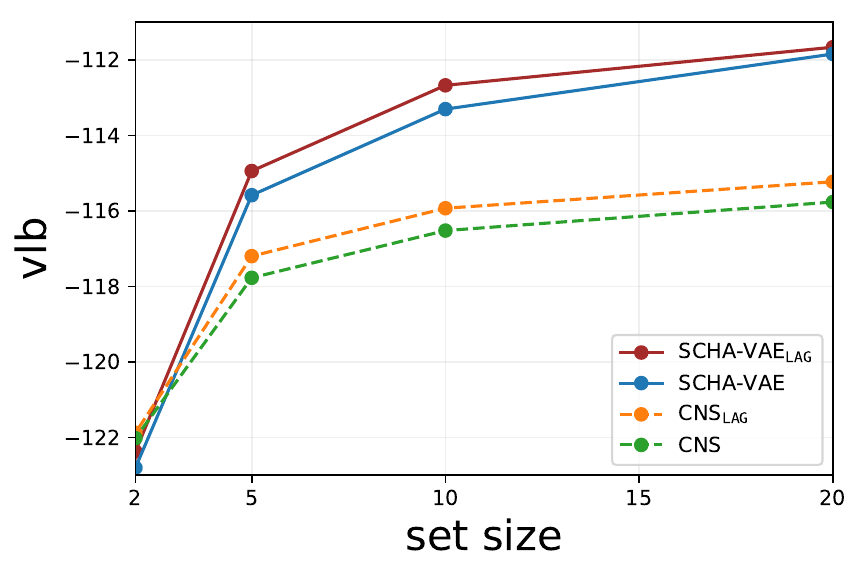}}
  \subfloat[Double-MNIST]{\includegraphics[width=.3\linewidth]{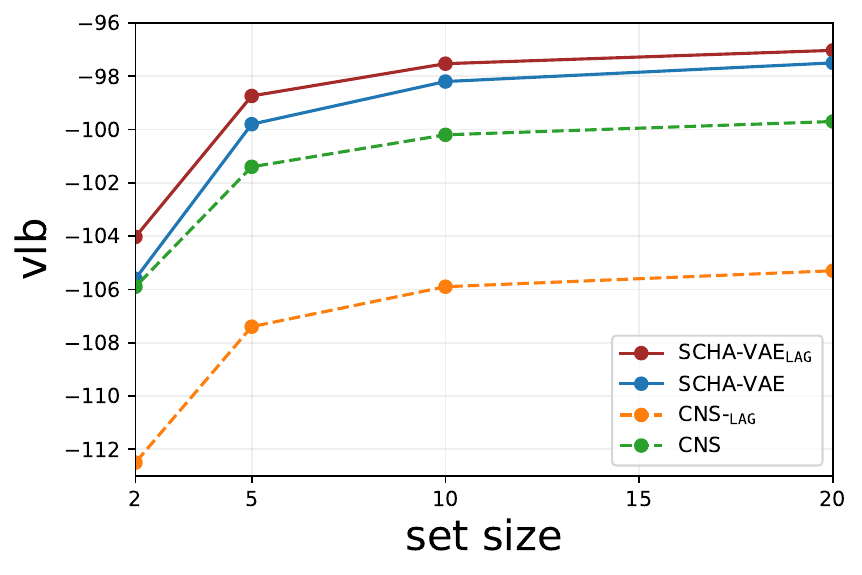}}
  \subfloat[Triple-MNIST]{\includegraphics[width=.3\linewidth]{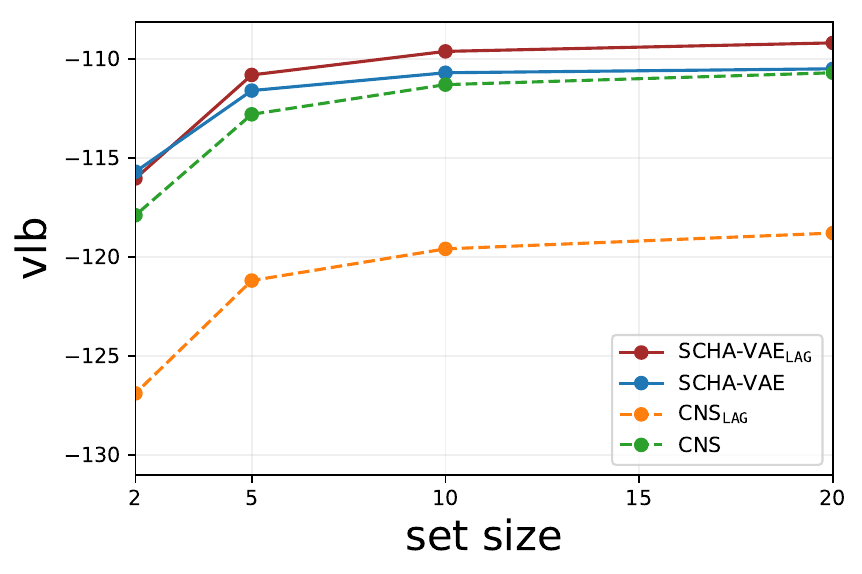}}
\caption{Transfer.
Model trained on Omniglot with set size 5 and tested on MNIST, DOUBLE-MNIST and TRIPLE-MNIST (from left to right) with different set size. We can see how our models perform better than a CNS. 
In particular SCHA-VAE with learnable aggregation (LAG) can adapt better to the new datasets. 
We test the model transfer capacities on scenarios of increasing complexity, using a subset of disjoint classes from simple out-distribution on MNIST and more challenging out-distribution generalization on variants with 20 and 200 classes.}
\label{fig:generative_metrics_transfer_cardinality}
\end{figure*}
In Appendix B, we report out of distribution classification performance for the three approaches described in Section \ref{sec:SCHA-VAE}. The models are trained on Omniglot for context size 5 and tested on MNIST also for context size 5 without adapting the distributions to the MNIST data.

\subsection{Hierarchy}
\begin{figure*}[ht]
  \centering
  \subfloat[Omniglot - Loss]{\includegraphics[width=.3\linewidth]{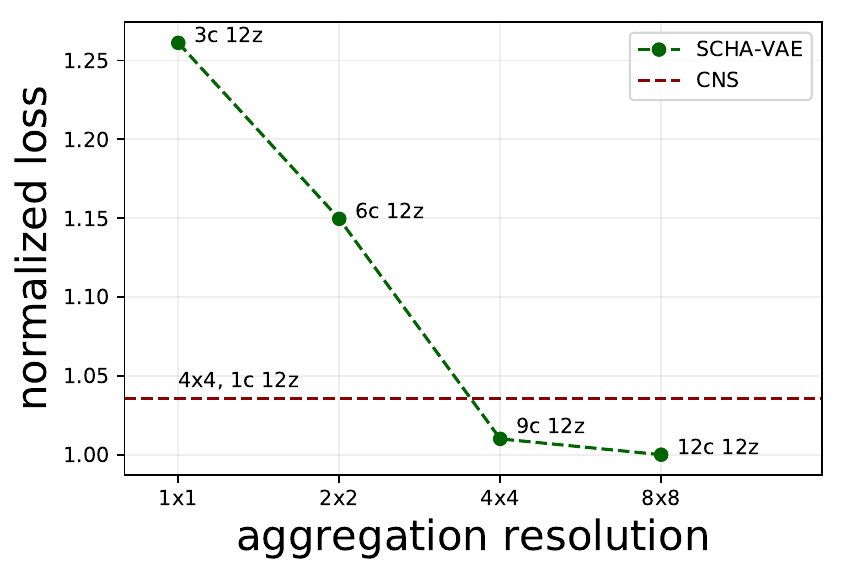}}
  \subfloat[CelebA, CIFAR100 - Loss]{\includegraphics[width=.3\linewidth]{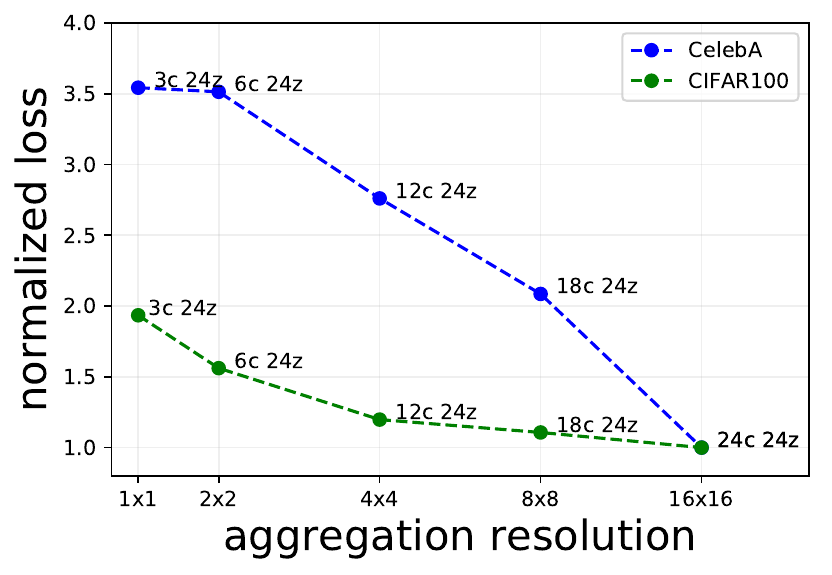}}
  \subfloat[CelebA, CIFAR100 - KL]{\includegraphics[width=.3\linewidth]{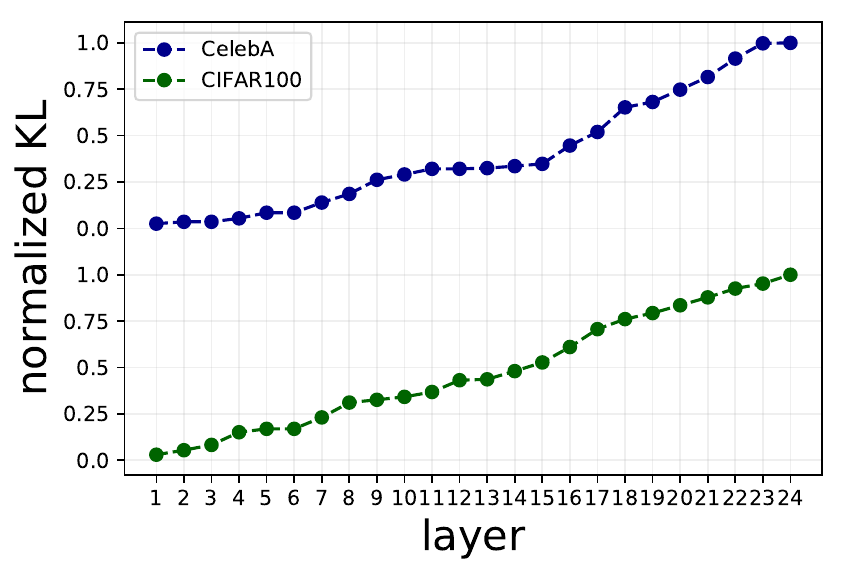}}
\caption{Role of depth and aggregation resolution in SCHA-VAE. 
Hierarchical context representation and multiresolution aggregation for SCHA-VAE on disjoint sets for Omniglot, CIFAR100 and CelebA. We normalize all the values to perform a simple comparison across models and datasets.
On the left and middle plots (a, b) we see that, reducing the prior depth and the aggregation resolution for c (without changing the depth for z) at inference time, the few-shot generation performance worsen, showing how SCHA-VAE relies on the hierarchy over c to extract better information from the conditioning set.
On the right we show how the cumulative KL increases with the stochastic layers, providing evidence that the layers at different resolutions are informative for the model.}
\label{fig:context_depth}
\end{figure*}

We claim that depth in the set representation and aggregation at multiple resolutions are essential to increase model expressivity. 
In Fig.~\ref{fig:context_depth} we provide empirical evidence. 
We train SCHA-VAE with 12 stochastic layers on Omniglot, and with 24 layers on CelebA and CIFAR100.
At inference, $c$ is bypassed for a certain number of layers, without modifying the depth for $z$, effectively decreasing the context depth, and we compute the test ELBO normalizing to one for easier comparison between datasets. 
If the hierarchy over c is redundant, removing a certain number of layers should not impact the overall result; if instead the model is using aggregation at different resolutions, then the result should be greatly impacted.
We see how removing the hierarchical information over $c$ worsen greatly the results for all the datasets, providing evidence that the set representation for few-shot generation.
On the right we plot the cumulative KL for CelebA and CIFAR100, showing how each layer contains additional information.
\section{Related Work}

\paragraph{Learning from Sets.}
In recent years a large corpus of work studied the problem of learning from sets~\cite{zaheer2017deep}, and more generally learning in exchangeable deep models~\cite{de20209, bloem2020probabilistic, bloem2019probabilistic}.
These models can be formulated in a variety of ways, but they all have in common a form of permutation invariant aggregation (or pooling mechanism) over the input set.
Deep Sets~\cite{zaheer2017deep} formalized the framework of exchangeable models. 
The Neural Statistician~\cite{edwards2016towards} was the first model proposing to learn from sets in the variational autoencoder framework and used a simple and effective mean pooling mechanism for aggregation. The authors explored the representation capacities of such model for clustering and few-shot supervised learning.
Generative Query Networks performs neural rendering~\cite{eslami2018neural} where the problem of pooling views arises. 
The Neural process family~\cite{garnelo2018neural, kim2019attentive}, where a set of point is used to learn a context set and solve downstream tasks like image completion and few-shot learning.
Set Transformers \cite{lee2019set} leverages attention to solve problems involving sets.
PointNet~\cite{qi2017pointnet} models point clouds as a set of points.
Graph Attention Networks~\cite{velickovic2017graph} aggregate information from related nodes using attention.
Associate Compression Network~\cite{graves2018acn} can be interpreted in this framework, where a prior for a VAE is learned using the top-knn retrieved in latent space.
In this work we build on ideas and intuitions in these works, with a focus on generative models for sets.
SetVAE~\cite{kim2021setvae} proposes a VAE for point-clouds, showing that processing the input set at multiple resolutions is a promising direction for set-based latent variable models.

\paragraph{Few-Shot Generative Models.}
Historically the machine learning community has focused its attention on supervised few-shot learning, solving a classification or regression task on new classes at test time given a small number of labeled examples.
The problem can be tackled using metric based approaches~\cite{vinyals2016matching, snell2017prototypical, oreshkin2018tadam}, 
gradient-based adaptation~\cite{finn2017model}, optimization~\cite{ravi2016optimization}.
More generally, the few-shot learning task can be recast as bayesian inference in hierarchical modelling~\cite{grant2018recasting, ravi2018amortized}. 
In such models, typically parameters or representation are conditioned on the task, and conditional predictors are learned for such task.
In~\cite{xu2019metafun} an iterative attention mechanism is used to learn a query-dependent task representation for supervised few-shot learning.
Modern few-shot generation in machine learning was introduced in~\cite{lake2015human}. 
The Neural Statistician~\cite{edwards2016towards} is one of the first few-shot learning models in the context of VAEs.
However the authors focused on downstream tasks and not on generative modeling.
The model has been improved further increasing expressivity for the conditional prior using powerful autoregressive models~\cite{hewitt2018variational}, a non-parametric formulation for the context~\cite{wu2020meta} and exploiting supervision~\cite{garnelo2018neural}.
~\cite{rezende2016one} proposed a recurrent and attentive sequential generative model for one-shot learning based on~\cite{gregor2015draw}.
Powerful autoregressive decoders and gradient-based adaptation are employed in~\cite{reed2017few} for one-shot generation. 
The context $\mathbf{c}$ in this model is a deterministic variable.
In GMN~\cite{bartunov2018few} a variational recurrent model learns a per-sample context-aware latent variable. 
Similar to our approach, GMN learns a flexible context, learning an attention based kernel that can handle generic datasets. 
However the context-aware representation scales quadratic with the input size, there is no separation between global and local information in latent space, and the input set is processed in an arbitrary autoregressive order, and not in a permutation invariant manner.
Finally, recent large-scale autoregressive language models~\cite{brown2020language} exhibit non-trivial few-shot capacities. 

\section{Conclusion}
Leveraging recent advances in deep latent variable models, we propose a new class of hierarchical latent variable models for few-shot generation.
We ground our formulation in hierarchical inference and a learnable non-parametric aggregation. 
We show how simple hierarchical inference is a viable adaptation strategy.
We perform extensive empirical evaluation in terms of generative metrics, sampling capacities and transfer properties. The proposed formulation is completely general and we expect there is large potential for improving performance by combining it with state of the art VAE architectures. Since benchmarks often come with grouping information, using a hierarchical formulation is a generic approach to improve generative capabilities.

\small
\section*{Acknowledgement}
We would like to thank
Anders Christensen,
Marco Ciccone,
Andrea Dittadi,
Pierluca D'Oro,
S{\o}ren Hauberg,
Valentin Liévin,
Didrik Nielsen,
Mathias Schreiner,
Timon Willi,
Max Wilson
for insightful comments and useful discussions.

\bibliographystyle{_ICML/icml2022}
\bibliography{biblio.bib}

\clearpage
\normalsize
\appendix
\onecolumn
\vspace{1cm}
\textbf{Code} : \url{https://github.com/georgosgeorgos/hierarchical-few-shot-generative-models} 

\vspace{1cm}
\textbf{Notation.}
\begin{tcolorbox}
\begin{itemize}
    \item \textbf{Bold} $\quad\quad\quad\quad\,\,\rightarrow$ \textbf{all layers}
    \item CAPITALIZED $\rightarrow$ ALL SAMPLES
\end{itemize}
\vspace{.3cm}
\begin{itemize}[leftmargin=5.5mm]
    \item[$\circ$] \textbf{Bold} CAPITALIZED: \textbf{all layers}, ALL SAMPLES $\mb{Z} = \{z^s_l \}^{S, L}_{s=1, l=1}$.
    
    \item[$\circ$] \textbf{Bold}: \quad\quad\quad \quad\quad\quad\,\,  \textbf{all layers}, one sample\quad\quad\,  $\mb{z}_s = \{z^s_l\}^L_{l=1}$.
    
    \item[$\circ$] CAPITALIZED: \quad\quad\, one layer, ALL SAMPLES $Z_l = \{z^s_l\}^{S}_{s=1}$.
\end{itemize}
\end{tcolorbox}

\begin{tcolorbox}
\begin{multicols}{2}
\begin{itemize}[topsep=0pt,parsep=0pt,partopsep=0pt,labelwidth=2cm,align=left,itemindent=2cm]
    \item[Top Prior $\mb{c}$:] $p_{\theta}(c_{L})$
    \item[Top Prior $\mb{z}$:] $p_{\theta}(z_{L} \vb c_{L})$
    \item[Prior $\mb{c}$:] $p_{\theta}(c_{l} \vb c_{l+1}, Z_{l+1})$
    \item[Prior $\mb{z}$:] $p_{\theta}(z_{l} \vb z_{l+1}, c_{l})$
    \item[Observation $\mb{x}$:] $p_{\theta}(x \vb z_{1:L}, c_{1:L})$
\end{itemize}

\columnbreak

\begin{itemize}[topsep=0pt,parsep=0pt,partopsep=0pt,labelwidth=2.5cm,align=left,itemindent=2cm]
    \item[Top Posterior $\mb{c}$:] $q_{\phi}(c_{L} \vb X)$
    \item[Top Posterior $\mb{z}$:] $q_{\phi}(z_{L} \vb c_{L}, x)$
    \item[Posterior $\mb{c}$:] $q_{\phi}(c_{l} \vb c_{l+1}, Z_{l+1}, X)$
    \item[Posterior $\mb{z}$:] $q_{\phi}(z_{l} \vb z_{l+1}, c_{l}, x)$
    \item[Set Representation:] $h_l$
\end{itemize}
\end{multicols}
\end{tcolorbox}

\vspace{1cm}
\begin{figure}[h]
    \centering
    \includegraphics[width=0.48\linewidth]{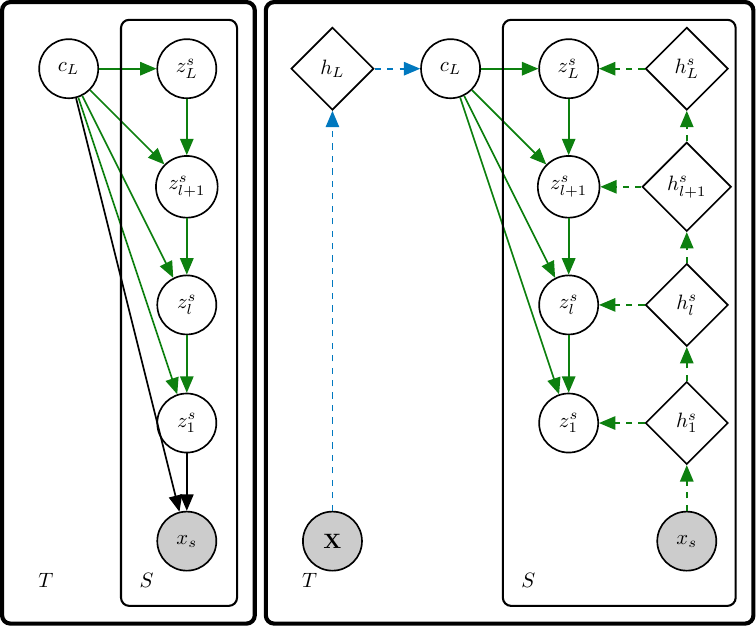}
    \includegraphics[width=0.48\linewidth]{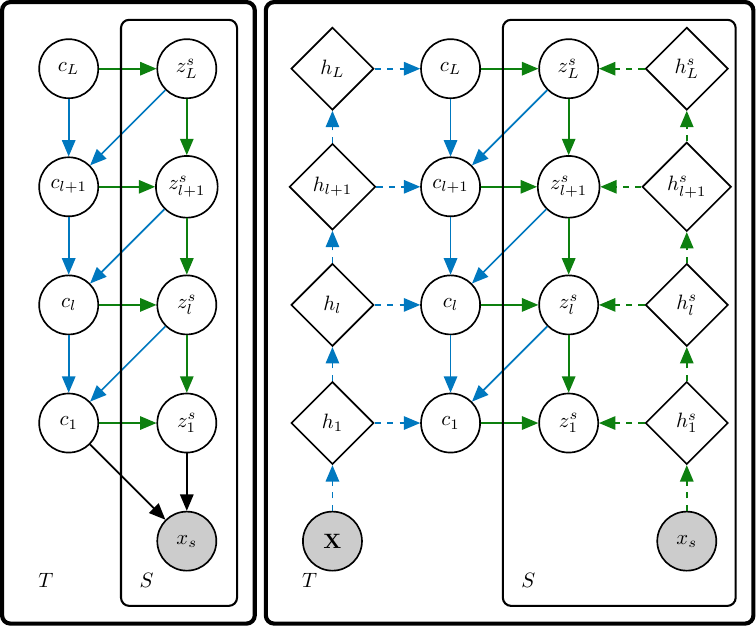}
    \caption{Generation and Inference for a Neural Statistician (left) and a Hierarchical Few-Shot Generative Model (right)}
\end{figure}

\clearpage
\newpage
\section{Model Derivation}
\label{appendixA}

In this section we derive the generative model, inference model and lower-bound for a Basic Neural Statistician (bNS), a Neural Statistician (with hierarchy over $z$, this is the model used as baseline in the paper) (NS), and a Hierarchical Few-Shot Generative Model (HFSGM) with hierarchy over $z$ and $c$. 
Doing so we can underline similarities and differences among the formulations.

\paragraph{Set Marginal.}
Our goal is to model a distribution $p(X) = \prod^S_{s=1} p(x_s)$ where $X  \{x_s\}^S_{s=1}$ is in general a small set ranging from 1 to 20 samples.
We sample these sets from a common process.
The Neural Statistician~\cite{edwards2016towards} introduces a global latent variable $c$ for a set $X$:
\begin{equation}
    p(X) 
    = \int p(X, c)~dc =\int p(c)~p(X | c)~dc 
    = \int p(c)~\prod^S_{s=1}p(x_s | c)~dc.
\end{equation}
Then we can introduce a per-sample latent variable $z$:
\begin{align}
\begin{split}
p(X) = 
\int p(c) 
\left[\int p(X,  Z |  c) dZ \right] dc &=
\int p(c) \prod_{s=1}^S 
\left[\int p(x_s,  z_s |  c) dz_s \right] dc \\
& = \int p(c) \prod_{s=1}^S 
\left[\int p(x_s | z_s, c) p(z_s | c) dz_s \right] dc,
\end{split}
\end{align}
where $X = \{x_s\}^S_{s=1}$ is a set of images, $c$ is a latent variable for the set, $Z = \{z_s \}^S_{s=1}$ are latent variable for the samples in the set.
The formula above is the basic marginal for all the NS-like model.

\subsection{Generative Model}
We can think of the model as composed of three components: a hierarchical prior over $z$, a hierarchical prior over $c$, and a render for $x$. 
For each hierarchical prior, the top distribution is not autoregressive and act as an unconditional prior for $c$ and a conditional prior for $z$.
In the following all the latent distributions are normal distributions with diagonal covariance. 
Given the hierarchical nature of our model, these assumptions are not restrictive, because from top to down in the hierarchy we build an expressive structured mean field approximation.
The decoder is Bernoulli distributed for binary datasets.
In the following equation we use 
$Z=\{z_s\}^S_{s=1}$, 
$\mb{Z}=\{\mb{z}_s\}^S_{s=1}$, 
$\mb{z}_s =\{z^s_l\}^L_{l=1}$, 
$\mb{c} = \{ c_l \}^L_{l=1}$.
Each of these equation can be written per-set or per-sample. 
We choose to write everything in a compact format writing per-set equations.

\paragraph{bNS.} In this setting both $z$ and $c$ are shallow latent variables:

\begin{equation}
p(X, Z, c) = 
p(X | Z, c)
p(c)
\Big[
p(Z | c)
\Big]
\end{equation}

\paragraph{NS.} $\mb{z}$ is a hierarchy:

\begin{equation}
p(X, \mb{Z}, c) = 
p(X | \mb{Z}, c)
p(c)
\left[
p(Z_L | c)
\prodl
p(Z_l | Z_{l+1}, c)
\right]
\end{equation}

\paragraph{HFSGM.} Both $\mb{z}$ and $\mb{c}$ are a hierarchy.
We increase the model flexibility. 

Using such hierarchy we can:
\begin{itemize}
    \item learn a structured mean field approximation for $c$; 
    \item jointly learn $c$ and $Z$, informing different stages of the learning process;
    \item incrementally improve $c$ through layers of inference.
\end{itemize}

\begin{align}
\begin{split}
p(X, \mb{Z}, \mb{c}) &= 
p(X | \mb{Z}, \mb{c})
p(c_L)
\left[
p(Z_L | c_L)
\prodl
p(Z_l, c_l | Z_{l+1}, c_{l+1})
\right] \\
p(Z_l, c_l | Z_{l+1}, c_{l+1}) &= p(Z_l | Z_{l+1}, c_l)~p(c_l | c_{l+1}, Z_{l+1}).
\end{split}
\end{align}

\subsection{Inference Model}
We learn the model using Amortized Variational Inference.
In a NS-like model, inference is intrinsically hierarchical: the model encodes global set-level information in $c$ using $q(c | X)$.

\paragraph{bNS.}
\begin{equation}
    q(c, Z | X) = q(c | X) \Big[q(Z | c, X)\Big]
\end{equation}

\paragraph{NS.}
\begin{equation}
    q(c, \mb{Z} | X) = q(c | X) \Big[q(\mb{Z} | c, X)\Big]
\end{equation}

\paragraph{HFSGM.}
\begin{align}
\begin{split}
q(\mb{c}, \mb{Z} | X) &= q(c_L | X) 
\left[ q(Z_L | c_L, X)
\prod^{L-1}_{l=1} q(Z_l, c_l | Z_{l+1}, c_{l+1}, X) 
\right] \\
q(Z_l, c_l | Z_{l+1}, c_{l+1}, X)  &= q(Z_l | Z_{l+1}, c_l, X)~q(c_l | c_{l+1}, Z_{l+1}, X).
\end{split}
\end{align}

\subsection{Lower-bound}
We learn the model by Amortized Variational Inference similarly to recent methods.
However we are in presence of two different latent variables, and we need to lower-bound wrt both.

\paragraph{bNS.}
\begin{align}
\begin{split}
    \log p(X) \geq~&\E_{q(c, Z | X)}\left[\log \dfrac{p(X, Z, c)}{q(c, Z| X)}\right] =\\
                  ~&\E_{q} \left[\sum^{S}_{s=1} \log p(x_s | z_s, c) \right] +
                  \E_q \left[ \sum^{S}_{s=1} \log \dfrac{p(z_s | c)}{q(z_s | c, x_s)}\right] 
                  - \KL(q(c | X), p(c)) = -\mathcal{L}(X).
\end{split}
\end{align}

\paragraph{NS.}
\begin{align}
\begin{split}
    \log p(X) \geq~&\E_{q(c, \mb{Z} | X)}\left[\log \dfrac{p(X, \mb{Z}, c)}{q(c, \mb{Z}| X)}\right]= \\
                ~&\E_{q} \left[\sum^{S}_{s=1} \log p(x_s | \mb{z}_s, c) \right] + 
                \E_q \left[\sum^{L-1}_{l=1} \sum^{S}_{s=1} \log \dfrac{p(z^s_l | z^s_{l+1}, c)}{q(z^s_l | z^s_{l+1}, c, x_s)}\right] + 
                \E_q \left[\sum^{S}_{s=1} \log \dfrac{p(z_L | c)}{q(z_L | c, x_s)}\right] \\
                -~&\KL(q(c | X), p(c)) = -\mathcal{L}(X).
\end{split}
\end{align}

\paragraph{HFSGM.}
\begin{align}
\begin{split}
    \log p(X) \geq~&\E_{q(\mb{c}, \mb{Z} | X)}\left[\log \dfrac{p(X, \mb{Z}, \mb{c})}{q(\mb{c}, \mb{Z}| X)}\right] =\\
    ~&\E_{q} \left[\sum^{S}_{s=1} \log p(x_s | \mb{z}_s, \mb{c}) \right] + 
    \E_q \left[\sum^{L-1}_{l=1} \sum^{S}_{s=1} \log \dfrac{p(z^s_l | z^s_{l+1}, c_{l})}{q(z^s_l | z^s_{l+1}, c_l, x_s)}\right] + 
    \E_q \left[\sum^{S}_{s=1} \log \dfrac{p(z_L | c_L)}{q(z_L | c_L, x_s)}\right] +\\
    ~&\E_q \left[\sum^{L-1}_{l=1} \log \dfrac{p(c_l | c_{l+1}, Z_{l+1})}{q(c_l | c_{l+1}, Z_{l+1}, X)}\right] 
    - \KL(q(c_L | X), p(c_L)) = -\mathcal{L}(X).
\end{split}
\end{align}

\subsection{Loss}
The final loss for all the models is computed per-sample.
Given a distribution of $T$ sets (or tasks) of size $S$, the training loss is: $L = \dfrac{1}{ST} \sum_T \mathcal{L}(X_t)$.

\subsection{Evaluation}
For VAEs we evaluate the models approximating the log marginal likelihood using $S$ importance samples:
\begin{equation}
    \mathrm{MLL}(x) = \log \dfrac{1}{IS}\sum^{IS}_{is=1} \dfrac{p(x, z_s)}{q(z_{is} | x)} \quad z_{is} \sim q(z | x).
\end{equation}
For hierarchical models like the NS we use:
\begin{equation}
    \mathrm{MLL}(X) = 
    \log \dfrac{1}{IS} \sum^{IS}_{is=1} \dfrac{p(X, Z_{is}, c_{is})}{q(Z_{is}, c_{is} | X)} \quad Z_{is} \sim q(Z | c, X), c_{is} \sim q(c | X). 
\end{equation}

\subsection{Learnable Aggregation}
In this subsection we describe more explicitly the aggregation mechanism.
Given a set $X$, embeddings for samples in the set $h_s = f_{\phi}(x_s)$, and aggregated statistics $r = 1/S \sum^{S}_{s=1} h_s$, we can compute attention weights for a NS$_{\mathtt{LAG}}$ as follows:
\begin{align}
\begin{split}
    \alpha(r, h_s) & = \sigma ( \texttt{dot}(q(r), k(h_s)) )\\
    r_{\mathtt{LAG}}        & = \sum^{S}_{s=1} \alpha(r, h_s)~v(h_s)   \\
    q_{\phi}(c ~|~ X) &= \mathcal{N}(c ~|~ \mu(r_{\mathtt{LAG}}),\sigma(r_{\mathtt{LAG}})),
\end{split}
\end{align}
where $q$, $k$ and $v$ are linear layers and $\textrm{sim}$ is the dot-product scaled by the square root of the representations dimensionality.
This approach is inspired by~\cite{lee2019set} and resembles self-attention with a fundamental difference: instead to map from $S$ samples to $S$ samples, we map from $S$ samples to a per-task learnable aggregation.
The query input is a handcrafted aggregation (mean, max pooling). 
We improve the query scoring the handcrafted aggregation with the samples in the set.

For a full SCHA-VAE$_{\mathtt{LAG}}$, we can similarly write:
\begin{align}
\begin{split}
    r & = \dfrac{1}{S} \sum^{S}_{s=1} f_{\phi}(x_s, z_s, c) \\
    \alpha(r, h_s, z_s, c) & = \sigma ( \texttt{dot}(q(r), k(h_s, z_s, c))) \\
    r_{\mathtt{LAG}}  & = \sum^{S}_{s=1} \alpha(r, h_s, z_s, c)~v(h_s, z_s, c)   \\
    q_{\phi}(c_{l} ~|~ c_{l+1}, Z_{l+1}, X) &= \mathcal{N}(c_{l} ~|~ \mu(r^l_{\mathtt{LAG}}),\sigma(r^l_{\mathtt{LAG}})),
\end{split}
\end{align}

\clearpage
\subsection{Sampling Algorithms}

\begin{algorithm}
\caption{Conditional and Refined Sampling SCHA-VAE}
\begin{multicols}{2}
\begin{algorithmic}
    \STATE {\bfseries Input:} $X$
    \STATE \textbf{single pass}
    \STATE $c_L \sim q(c_L|X)$
    \STATE $\mb{z},\mb{c}_{<L}\sim p(\mb{z},\mb{c}_{<L}|c_L)$
    \STATE $x\sim p( x|\mb{z},\mb{c})$
    \RETURN $x$
\end{algorithmic}
\columnbreak
\begin{algorithmic}   
    \STATE {\bfseries Input:} $X$
    \STATE \textbf{single pass}
    \STATE $c_L \sim q(c_L|X)$
    \STATE $\mb{z},\mb{c}_{<L}\sim p(\mb{z},\mb{c}_{<L}|c_L)$
    \STATE $x\sim p( x|\mb{z},\mb{c})$
    \REPEAT
    \STATE $\tilde{X}=[X,x]$
    \STATE $\tilde{Z},\mb{c} \sim q(\tilde{Z},\mb{c}|\tilde{X})$
    \STATE$ \tilde{X}' \sim p(\tilde{X}'|\tilde{Z},\mb{c})$
    \STATE $x = x'$
    \UNTIL{$\texttt{Convergence}$}
    \RETURN $x$
 \end{algorithmic}
\end{multicols}
\end{algorithm}

\clearpage
\newpage
\section{Additional Experiments}
\label{appendixB}
\subsection{Generalization}

\begin{figure}[ht]
	\begin{minipage}{.55\columnwidth}
	\centering
	\caption{Lowerbound in Bits per Dimension (bpd) for CelebA 64x64 test set (unknown classes).
	We train a CNS, CNS$_{\texttt{LAG}}$, and SCHA-VAE with 3 stochastic layers and process the set information at one single resolution (4x4). We notice that even in this unfavourable scenario, where SCHA-VAE cannot exploit depth and multiresolution aggregation, we gain in terms of likelihood thanks to the hierarchical representation for c.}
		\begin{tabular}{lcccc}
		\toprule
		& \multicolumn{4}{c}{CelebA} \\
		& Depth c & Res $c$ & Depth $z$ & NLL (bpd) \\
		\midrule
		CNS & 1 & 4x4 & 3 & $\leq$ 4.2235   \\
		CNS$_{\mathtt{LAG}}$  & 1 & 4x4 & 3 & $\leq$ 4.2126   \\
		SCHA-VAE     & 3 & 4x4 & 3 & $\leq$ \textbf{4.1552} \\
		 \bottomrule
		\end{tabular}
	\end{minipage}%
	\begin{minipage}{.45\columnwidth}
		\centering
		\includegraphics[width=.9\columnwidth]{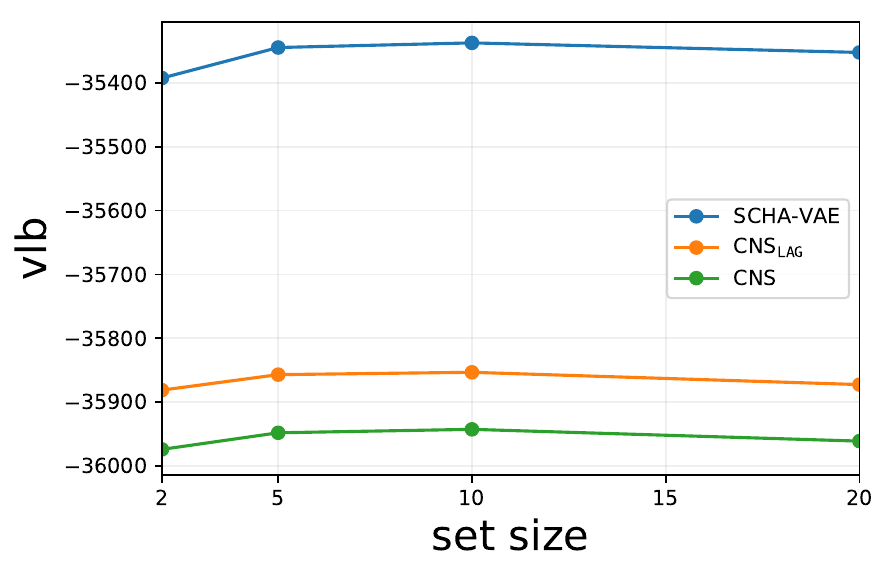}
		\caption{Set Cardinality.
    All models are trained with homogeneous (one concept) sets.
    Lower-bounds for models trained with input set size 5 varying the test set cardinality from 1 to 20 on CelebA. All the models struggle with set size larger than 10 because of the large variety in each set (age, perspective, general look).
We can see that learnable aggregation and hierarchical inference both help in modeling the dataset.}
		\label{fig:generative_metrics_omniglot_cardinality}
	\end{minipage}
\end{figure}

\begin{table}[ht]
	\centering
	\caption{Omniglot test negative loglikelihood for convolutional and autoregressive based models. We train a deeper SCHA-VAE with 24/33 stochastic layers and input set size 5. 
	We augment the sets with small translations and rotations.
	We approximate the likelihood computing the importance weighted bound with S=1000 on homogeneous sets of cardinality 20 (all the samples for a given class/character).
	The Variational Homoencoder (VHE) uses a PixelCNN to encode $c$ and decode $x$. 
	Generative Matching Network (GMN) uses RNN layers for encoding/decoding, and attention for aggregation. 
	We show the complexity to compute $c$, forward $p(X | c)$ and sample $p(x|X)$.
	We can see that SCHA-VAE is competitive with and without autoregressive components.
	$n$ : set input dimensionality. 
	$d$: image dimension, where in few-shot generation $d >> n$.
	}
		\begin{tabular}{lcccc}
		\toprule
		       & Aggregation & Sampling & NLL (nats)  \\
		\midrule
		\emph{without autoregressive components} \\
		VHE~\cite{hewitt2018variational} & $O(n)$  & $O(1)$ & 104.67\\
		NS~\cite{edwards2016towards}     & $O(n)$  & $O(1)$ & 102.84\\
		CNS                              & $O(n)$  & $O(1)$ & 77.21\\
		SCHA-VAE (Ours)                  & $O(n)$  & $O(1)$ & \textbf{64.34}\\
		\midrule
		\emph{with autoregressive components} \\
		NS + PixelCNN                    & $O(n)$   & $O(d)$   & 73.50\\
		GMN~\cite{bartunov2018few}       & $O(n^2)$ & $O(n)$   & 62.42\\
		VHE~\cite{hewitt2018variational} & $O(n)$   & $O(d)$ & \textbf{61.22}\\
		\bottomrule
		\end{tabular}
		\label{tab:autoregressive}
\end{table}

\clearpage
\subsection{Sampling}
\begin{figure}[ht]
    \centering
     \includegraphics[width=.9\linewidth]{img/experiments/sampling/sampling-refine-omni-ctns.pdf}
    \includegraphics[width=.9\linewidth]{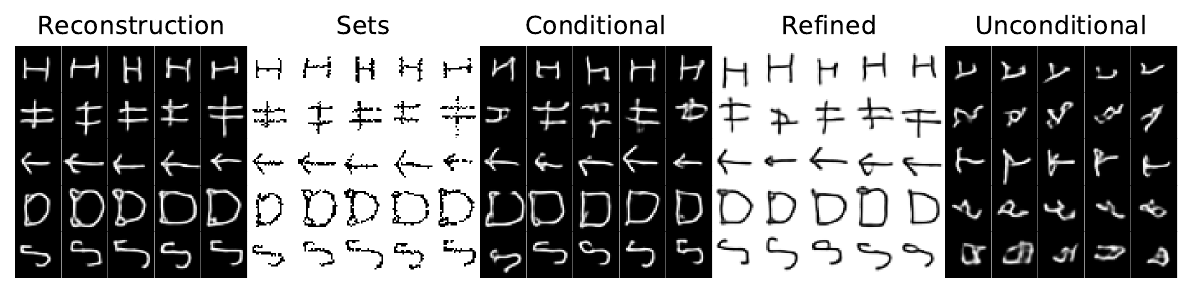}
    \includegraphics[width=.9\linewidth]{img/experiments/sampling/sampling-chfsgm_multi_celeba.pdf}
    \caption{Sampling. Different ways to sample the model.
    (Top): Refined samples obtained using Algorithm~\ref{alg:sampling}. 
    Given a small set from an unknown character (right on black background), we sample the model and then refine iteratively using the inference model. We show 20 iterations from left to right. 
    We can see how the generative process refines its guess at each iteration improving $\mathbf{c}$ and $z$ in a joint manner. \\
    (Middle):
    Stochastic reconstruction, input sets, conditional sampling, Refined sampling and unconditional sampling (sometimes referred to as imagination) on Omniglot.
    The models are trained on subsets of Omniglot and tested on disjoint characters. \\
    (Bottom):
    Stochastic reconstruction, input sets, conditional sampling, and unconditional sampling (sometimes referred to as imagination) on CelebA.
    The models are trained on subsets of CelebA and tested on disjoint identities.
    }
    \label{fig:sampling_omniglot}
\end{figure}

\clearpage
\subsection{Transfer}

\begin{figure}[ht]
  \centering
  \subfloat[MNIST]{\includegraphics[width=.45\linewidth]{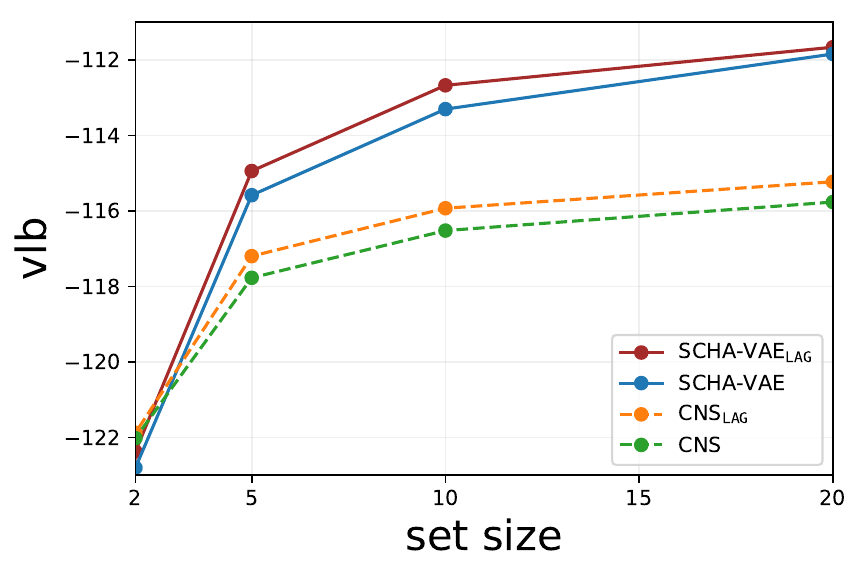}}
  
  \subfloat[Double-MNIST]{\includegraphics[width=.45\linewidth]{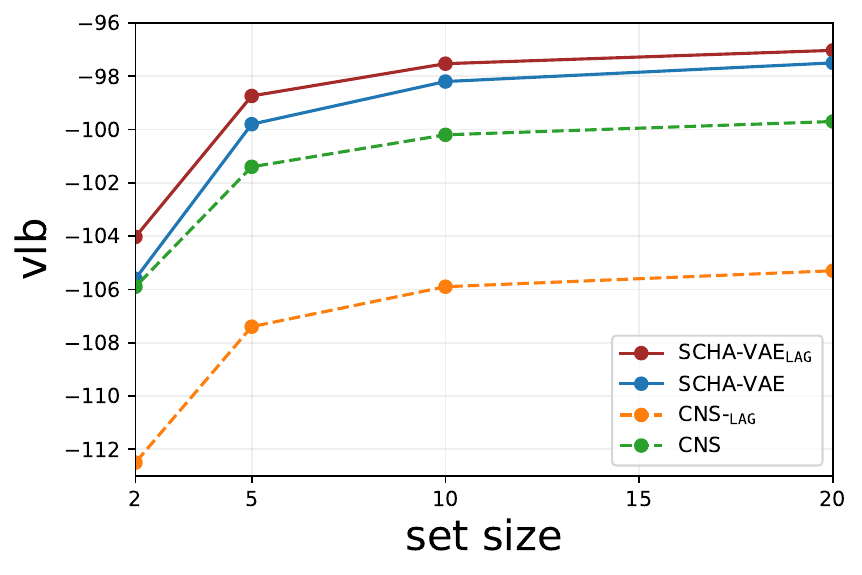}}
  \quad
  \subfloat[Triple-MNIST]{\includegraphics[width=.45\linewidth]{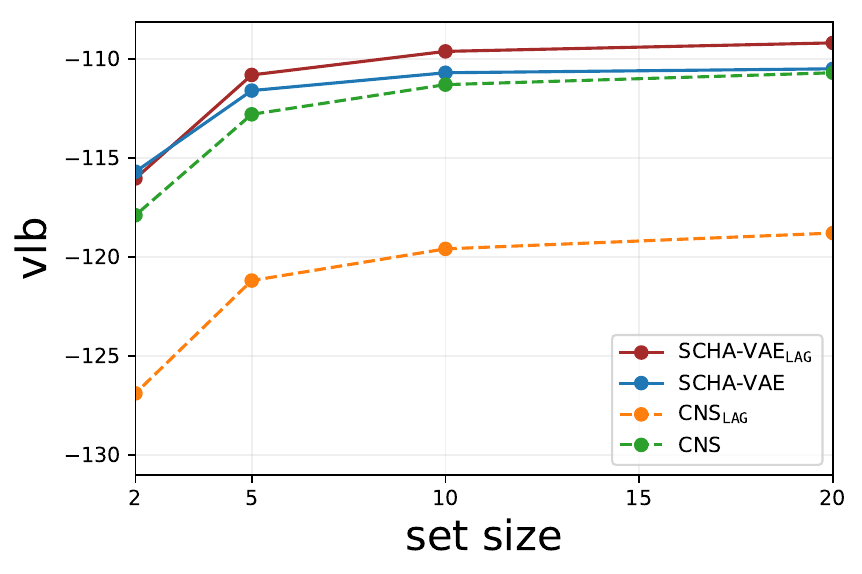}}
\caption{Transfer.
Model trained on Omniglot with set size 5 and tested on MNIST, DOUBLE-MNIST and TRIPLE-MNIST (from left to right) with different set size. We can see how our models perform better than a CNS. 
In particular SCHA-VAE with learnable aggregation (LAG) can adapt better to the new datasets. 
We test the model transfer capacities on scenarios of increasing complexity, using a subset of disjoint classes from simple out-distribution on MNIST and more challenging out-distribution generalization on variants with 20 and 200 classes.}
\label{fig:generative_metrics_transfer_cardinality}
\end{figure}

\clearpage
\subsection{Classification}
In this paper our main interest is in improving few-shot generation specifically through the lens of hierarchical inference, multiresolution and learnable aggregation.
However is natural to ask how the model perform on downstream tasks, in particular on the supervised few-shot learning task.
Here we focus on a simple experiment: we train the models on Omniglot and test on MNIST without any form of adaptation.
We consider different ways to approximate a classifier.
We can use the fitted generative model as part of a few-shot Bayes classifier:
$$
    p(y|x,X) = \frac{p(y)p(x|X_y)}{\sum_{y'} p(y')p(x|X_{y'})}, %
$$
where $X=\{X_1,\ldots,X_C\}$ is the set of datasets for the $C$ classes. 
In Appendix B we investigate two approaches that approximate the predictive distribution $p(x|X_y)$ and one based on $q(c | X)$:
\begin{itemize}
    \item ELBO difference:
$$
\log p(x|X_y)  = \log p(x,X_y) - \log p(X_y) 
    \approx \mathrm{ELBO} (x,X_y) - \mathrm{ELBO} (X_y).
$$ 
\item Equation (6): 
Sample $q(c | X_y)$ and evaluate: $$\E_{q(c)}[\log p(x | c)].$$
\item The classification approach of~\cite{edwards2016towards}: 
$$\mathop{\mathrm{argmax}}_y \KL(q(c | X_y), q(c|x)).$$ 
\end{itemize}
\vspace{.3cm}
\begin{table}[h]
\begin{center}
\caption{Few-shot classification.
Models trained on Omniglot and tested on binarized MNIST. 
Input set dimension 5.
For consistency with the KL classifier used in NS, 
we use only one layer of posterior, the one closer to the data.}

\begin{tabular}{lcccc } 
\toprule
& AGG  & $\mathbb{ELBO}\left[x | X\right]$ & $\mathbb{KL}\left[q_{X}, q_{x}\right]$ & $\mathbb{E}_q \left[\log p(x | c)\right]$ \\
\midrule
NS     & MEAN & 0.46 & 0.73 & 0.74  \\
CNS    & MEAN & 0.41 & \textbf{0.76} & \textbf{0.76}  \\ 
CNS    & LAG  & 0.33 & 0.75 & \textbf{0.76}  \\ 
SCHA-VAE & MEAN & \textbf{0.52} & 0.71 & 0.75  \\
\bottomrule
\end{tabular}
\end{center}
\label{fig:few-shot-cls}
\end{table}

\clearpage
\newpage
\section{Implementation Details}
\label{appendixC}
The baseline models are close approximation of the Neural Statistician adapted from: \url{https://github.com/conormdurkan/neural-statistician}.
The images are encoded using a shared encoder with 3x3 convolutions plus batch normalization. resolution is halved using stride 2. 
The decoder is the same, with resolution doubled using transposed convolutions.
More powerful and expressive decoders can be employed~\cite{oord2016pixel, salimans2017pixelcnn++} or multi-resolution deep latent variable models~\cite{maaloe2019biva, vahdat2020NVAE, child2020very}.
We do not use sample dropout (removing random samples from the input set and use the set statistics as additional features).

The loss is a weighted negative lower-bound: $L = -\mathrm{VLB}(\alpha) =  (1 + \alpha) \mathrm{REC} + (\mathrm{KL}_z + \mathrm{KL}_c) / (1 + \alpha)$ where alpha is annealed decreasing at each epoch $\alpha = \alpha * \alpha_{\mathtt{step}}$, with $\alpha_{\mathtt{step}} < 1$ at the beginning of training. 
This re-weighting tends to magnify the importance of the likelihood term and reduce the risk of posterior collapse at the beginning of training. 
Learning is slower but typically the model and posterior learned are better.

\begin{table}
\begin{center}
\caption{Relevant Hyperparameters for SCHA-VAE.
DMoL: Discretized Mixture Logistics.
VLB: Variation Lowerbound.}
\begin{tabular}{lccc} 
\toprule
& Omniglot & CelebA & CIFAR100 \\ 
\midrule
Dimension               &  1x28x28   &  3x64x64       &  3x32x32 \\
Number classes          &  1623      &  6349          & 100      \\
Classes Train           &  1000      &  4444          & 60       \\
Classes Val             &  200       &  635           & 20       \\
Classes Test            &  423       &  1270          & 20       \\
\midrule
$\alpha$                &  1$\div$2  &  1$\div$2      &  1$\div$2      \\
$\alpha$ step           &  0.5$\div$0.98 &  0.5$\div$0.98 &  0.5$\div$0.98 \\
Batch norm              &  \xmark    &  \xmark        & \xmark             \\ 
Batch size              &  128       &  16            & 32                 \\
Channels latent $c$     &  32        &  32            & 32                 \\
Channels latent space   &  128       &  128$\div$256  & 128                \\
Channels latent $z$     &  32        &  32            & 32                 \\
Classes per set         &  1         &  1             &  1                 \\
Epochs                  &  1000      &  600           & 600                \\ 
Heads                   &  4         &   4            & $-$                \\
Input set size          &  2$-$20 &  2$-$20     &  2$-$20         \\
Learning rate           &  $1e^{-3}$/$2e^{-4}$     &  $2e^{-4}$     & $2e^{-4}$ \\
Schedule                &  plateau/step   &  plateau  & plateau            \\
Likelihood              &  Bernoulli &  DMoL/Bernoulli &  DMoL/Bernoulli   \\
Loss                    &  VLB       &  VLB           & VLB                \\
Optimizer               &  Adam      &  Adam          &  Adam              \\
Residual layers         &  3         &  3             &  3                 \\
Resolution latent space &  1,2,4,8   &  1,2,4,8,16    &  1,2,4,8,16        \\
Stochastic layers       &  3$-$12 &  3$-$24     & 3$-$24          \\ 
Weight decay            & 0.00001    &  0.00001       &  0.00001           \\
\bottomrule
\end{tabular}
\end{center}
\end{table}

\paragraph{Modulation.}
The way we condition the prior and generative model greatly enhances the adaptation capacities of the model.
Other than conditioning the representations through direct concatenation or summation, we can condition directly the features and activations in the residual blocks.
FiLM~\cite{perez2017film} is a modulation module used in transfer learning and large-scale class conditional generation.
Given an input $c$ and a feature map $f$ with $H$ channels, FiLM learns an affine transformation with $2H$ modulation parameters $(\gamma_h, \beta_h)^H_{h=1}$ for each input $c$:
$$
    \textrm{FiLM}(z, c) = \gamma(c) f(z) + \beta(c).
$$
This affine transformation can be applied in different part of the prior and generative model.
In the main paper we use a simplification of such approach, applying $c$ to $z$ using only the bias term, i.e. $\gamma(c) = I$ and $\beta(c) = c$. 
Empirically we found that for tasks with a unique concept, like sets of characters and faces, this approach is effective.

\end{document}